# A Novel Convolutional Neural Network-Based Framework for Complex Multiclass Brassica Seed Classification

Elhoucine Elfatimi[a,*], Recep Eryiğit[b] and Lahcen Elfatim[c]

[a] *Department of pathology , University of Irvine, California, USA*
[b] *Department of Computer Engineering, Ankara University, Ankara 06830, Türkiye*
[b] *Computer Engineering Department, Polytechnic School of Montreal, 2500, Montreal, H3T 1J, Canada*

ABSTRACT

Agriculture research has accelerated during the last few years, but farmers still don't have adequate time or money for on-farm research because the bulk of their time is spent growing crops and running a farm. Seed classification can provide additional knowledge about quality production, seed quality control, and impurity identification. Early classification of seeds is critical for reducing the cost and risk of target field emergence, which can result in lost crop yield or even damage to downstream processes, such as harvesting. Seed sampling can aid growers in monitoring and controlling seed quality. It also increases precision for determining the type and level of seed purity, informs the farmer about any necessary adjustments to their management practices, and can improve the accuracy of yield estimation. This study proposed a new model based on CNN, which provides an efficient method for identifying and classifying Brassica seeds. The proposed system includes a model that uses advanced deep-learning techniques to classify ten common classes of Brassica seeds. This pioneering approach is thoughtfully crafted to tackle the inherent challenge posed by texture similarity in images. Furthermore, we evaluated the efficacy of our CNN model, compared the results to those of some pre-trained state-of-the-art architectures, and extracted the features for the best classification by changing the number and properties of layers in the image. The goal was to evaluate the feasibility of determining the architecture of the highest-performing model and the best training settings for these problems. Finally, the developed model is experimented with using our collected Brassica seed dataset. The results revealed that the designed model architecture generated a high accuracy value of 93%.

## 1. Introduction

In recent years, image analysis of seeds has emerged as a critical tool for biodiversity conservation and agricultural efficiency. The challenge of identifying and classifying plant species, particularly Brassica seeds, lies in the texture similarities and morphological variations among different types. This study introduces an innovative convolutional neural network (CNN) architecture specifically designed to address these challenges in Brassica seed classification. Unlike traditional methods, our approach leverages multiple convolutional layers with varying filter sizes and spatial dimensions to capture intricate and complex features from the input images. This novel architecture not only enhances the accuracy and robustness of seed classification but also provides practical solutions for real-world agricultural applications.

Image analysis of seeds has become critical for biodiversity conservation. As a result, identifying and classifying plant species on the planet is currently a major challenge. In order to tackle this challenge, many researchers have initiated and invested in a number of projects involving image processing and deep learning. Most of these projects face the same problem with the some methods, which necessitates working on different projects and employing various new technologies, each with its own set of advantages and disadvantages. Though completing all of the various challenging projects can be time-consuming and tedious, it is extremely beneficial in the end because it contributes to the creation of a system that can be easily maintained, improved, and expanded, especially as new models and algorithms are developed. Therefore, this type of system will continue to be used as a tool for more accurate seed classification and, eventually, will address the full set of image classificaiton challenges.

Several factors make classifiying and identifying brassica seeds images difficult: i.) Brassica variants are recognized to cross, resulting in seeds with intermediate characteristics; ii.) Dataset images may be very closely related and thus have very similar seeds (for example, Brassica Napus Var Annua, Brassica Napus Var Oleifera, Brassica Rapa Oleifera, Brassica Oleifera Var Gongylodes . . . ); iii.) Environmental factors can impact seed appear-ance; iv.) Immature seeds do not always display obvious characteristics; they may differ in texture and color from mature seeds. Therfore, creating new models and algorithim will not only contribute to a better understanding of fundamental seed traits and the interactions that control these traits, but it will also improve our capacity to produce more efficient and accurate classifications in other areas of image classification and the challenges they face. In this study, we focused primarily on developing and proposing a new CNN architecture for Brassica seed image classification. Furthermore, we assessed the effectiveness of our CNN model and extracted the features for the best classification by varying the number and properties of image layers. The goal was to find the architecture of the best performing model and the best training settings for these problems, which we then applied to the classification of Brassica seed images problems. We evaluated the effects of the proposed architectures and training settings on improving performance using different measurements. Therefore, this paper presents the complete research process of creating and implementing a new convolutional neural network (CNN) model for image classification in higher dimensional spaces, especially to classify complex Brassica seed images into 10 classes. The goal of this study is also to create a new Brassica dataset that did not previously exist and to evaluate the performance of our CNN architectures on this dataset. The images of the seeds were taken with a digital microscope in daylight at 1600 x 1200 pixels at 96 dpi. A dataset was created by randomly selecting a large number of images of each type. This dataset was created in order to evaluate our new CNN architectures and analyze the accuracy achieved, which will contribute to a better understanding of the applications of CNN models in agricultural data classification tasks.

*Corresponding author.
E-mail address(es): elhoucine.elfatimi123@gmail.com (E. Elfatimi).

The contribution of this work are through the following tasks:

1. We present a novel CNN model designed for the precise identification and classification of complex multiclass Brassica seeds, leveraging cutting-edge deep learning techniques. This innovative approach is meticulously crafted to address the inherent challenge of texture similarity in images, offering practical and tailored solutions to benefit the agricultural community.
2. Collecting and preprocessing a unique and complex dataset, followed by implementing the proposed framework for analysis and evaluation.
3. Imposing meticulous finesse by fine-tuning our model, and meticulously adjusting parameters. This precision ensures the model's consistent and optimal performance across diverse datasets.
4. Rigorously scrutinizing the proposed method's performance via a comprehensive array of performance metrics, meticulously benchmarking it against pre-trained state-of-the-art deep learning models. This comparative analysis is of profound significance, especially when considering practical applications in agriculture.

The developed model was experimented with using a collected Brassica seed dataset using different evaluation measures. This approach helped us to find and use the best-performing architectures and training settings, which had the highest accuracy in predicting class labels. It allows us to tune the training model and find a combination of architecture parameters, network topology, and weight settings with the best performance in predicting class labels.The models can be applied to other problems with similar characteristics.

The remainder of the paper is organized as follows: Section 2 provides a literature review on recent research on seed image classification. Section 3 describes the training process, the proposed model architecture, and the dataset used. Section 4 presents the experimental setup, performance evaluation, and results obtained, as well as a detailed description of the model architecture and comparison with some pretrained model. Finally, Section 5 concludes the paper and suggests future research.

## 2. Existing Work

Deep learning techniques, particularly CNNs, have revolutionized various fields, including agriculture, bringing greater flexibility, high efficiency, precision, accuracy, and cost-effective solutions to various problems faced by farmers in developing countries, especially in disease control, image classification, and experimental operations. For example, the classification control system of seeds such as Brassica seeds, which is important in most countries, including Turkey, can be considered a typical case. Previous systems for seed classification were developed in the early 1990s [1]. At that time, seed classification was done by people directly from images, which was time-consuming and unsuitable for high-throughput work. However, advanced deep learning techniques, especially CNN algorithms, can now accurately classify seeds based on visual inspections and provide a clear comparative advantage over these systems. As a result, CNN techniques are now extensively used in primary industries, including manufacturing and healthcare, as well as by farmers and agricultural scientists. Furthermore, several studies on developing and deploying Deep Learning techniques for seed classification have yielded positive results.

For instance, Agrawal et al. [2] carried out a comparative study on ML algorithms with the goal of classifying various grain seeds using Linear Discriminant Analysis (LDA), logistic regression (LR), a decision tree classifier (CART), k-Nearest Neighbors classifier (kNN), support vector machine (SVM), and Gaussian Nave Bayes (NB). The performance rates for both linear and non-linear algorithms were discussed in this study. The accuracy percentages for these six algorithms were just as follows: kNN 87.5%, LDA 95.8%, NB 88.05%, LR 91.6%, SVM 88.71%, and CART 88%; based on these results, it is clear that LDA consistently outperformed. In another study, Foysal et al.[3] used deep convolutional neural networks to classify images of healthy and uhealthy tomato leaves; the developed model reported an accuracy of 76%.

Gulzar et al. [4] proposed a system for seed classification based on Convolution Neural Networks. The proposed system includes a model that classifies 14 seeds, and the methods used in this study reported an accuracy of 99% for the test set and 99% for the training set. The CNN technique also was used by Keya et al.[5] to identify rice, sweet squash, corn, and gourd seeds. They provided a dataset of 1250 images; the training accuracy in this work ranged from 87% to 89% as a result.

Ali et al. [6] developed a method for classifying corn seeds using machine learning techniques that include the Bayes network (BN), random forest (RF), LogitBoost (LB), and MLP. The applied approaches were conducted well, with the RF method achieving 97.22% accuracy, the BN method with 97.67% accuracy, the LB method with 97.78% accuracy, and the MLP method with 98.93% accuracy.

Salimi et al. [7] used multispectral imaging (MSI) to classify five different damaged seed types in sugar beet seeds. With an accuracy rate of 82%, the classification model, which is based on MSI information, classified five different damaged seeds.

Minah et al.[8] conducted research to classify Brassica rapa varieties. They created three types of the phenotypic image from 156 Brassica rapa core collections to build AI-based classification models, and classification was performed using four different con-volutional neural network architectures. The result displays an accuracy of more than 87.72%.

With the help of 45 morphological features, Dubey et al. [9] were able to distinguish between three types of wheat, with an accuracy varying from 84% to 94% for each type. Hernandez et al. [10] also used color, statistical, and morphological features to classify barley and wheat seed grains, with an overall accuracy of 99%. Similarly, Shahid et al. [11] combined three distinct feature selection methods to identify the five types of wheat, he achieved accuracy close to 95% (ANN).

Loddo et al.[12] proposed A novel deep-learning technique for classifying and retrieving seed images. The researchers examined the classification performance of ten different CNN architectures and a new CNN model called SeedNet for seed image classification. For both datasets used, the results were more than 95% accurate.

There have been few studies that used CNN to classify seed varieties. For example, Maeda-Gutiérrez et al.[13] compared CNN-based architectures such as GoogleNet[14], AlexNet [15], Inception V3 [16], and Residual Network (ResNet 18 and 50)[17]. Their data set contained only one type of tomato seeds. In our study, we proposed an efficient model for seed classification based on CNN, a deep learning model with a high precision level in image feature extraction. Unlike most relevant studies, the dataset for this study contains ten different types of Brassica seeds.

According to the literature, deep learning techniques with different approaches are efficient for seed classification. However, studies of a few critical crops, particularly those using Brassica seed, were not found, and most existing studies face the same problem with some techniques, necessitating working on multiple projects and the use of diverse new technologies. As a result of the lack of this type of study on Brassica seeds, as well as the widespread use of these seeds, we focused more on Brassica seed image classification to increase crop yield and quality. We specifically classified seeds from ten types of Brassica (Brassica Napus Var Annua, Brassica Napus Var Oleifera, Brassica Nigra, Brassica Oleracea Gongylodes, BrassicaO leraceaLCAV rubra, Brassica Oleracea Rapa Brassica, Brassica Oleracea Var Gongylodes, and Brassica Rapa subsp. rapa) using a new model based on a convolutional neural network classifier, which is a modern method for computer recognition that has significantly improved efficiency in classifying Brassica seed types based on their images. Therefore, the model we propose is based on a convolutional neural network classifier, where different sizes of filters are used to decrease overfitting and increase accuracy. The model filters are used to extract a feature from the data, where each filter is generated using specific sizes and shapes in different positions. The analysis of existing approaches employed for seed classification is presented in Table 1.

In this study, the proposed model can learn more complex features because there are more convolution layers in the stack with smaller filter sizes than in previous studies, such as those using AlexNet or Google Net. Because our model is optimized to be small and efficient while sacrificing accuracy, which makes the model more powerful and easier to train.

In [8], the authors created AI models for classifying Brassica rapa varieties using four convolutional neural network architectures. They reported a lower accuracy of 87.72% than in the current study. The present study's performance achieved an average accuracy of 93%, proving the effectiveness of the created model. Using this approach, we could generate a variety of convolutional neural networks with higher accuracies than many previous studies, which either used different datasets or did not perform any cross-validation and fine-tuning. In our research, features from various layers, including fully connected layers, were analyzed, and the results of this study were used to improve outcomes in some cases, which is better than in previous studies. Furthermore, as we can see from the results in [5, 7, 9], the model created in this study is more accurate than in previous studies. In addition to these advantages, the current study has another advantage compared with previous studies that it can be applied in a wide variety of conditions, as the data of these studies were more complex and represented many aspects of the real world, which made it possible to transfer it from a large amount of data from the real world to train, showing a satisfactory level even in scenarios with a large number of variables and conditions, which led to a better generalization of the model and showed higher effectiveness on the related data, and this is exactly the result that researchers want from any model. The performance results of this study will be discussed in the results and discussion section of the paper.

## 3. Research Materials and Methods

This section presented a detailed description of the proposed model as well as the datasets that were used. The section also describes the steps taken to improve the performance of the proposed method implementation and finally presents the proposed CNN model architecture.

### 3.1. Dataset and Training Process.

In this study, a new and unique dataset that does not exist in the literature has been created, including ten different types of Brassica seeds. The prepared seeds classes were (Brassica Napus Var Annua, Brassica Napus Var Oleifera, Brassica Nigra, Brassica Oleracea Gongylodes, BrassicaO leraceaLCAV rubra, Brassica Oleracea Rapa Brassica, Brassica Oleracea Var Gongylodes, and Brassica Rapa subsp. rapa). Table 2 depicts the dataset's frequency distribution.

It is necessary to note that the number of images captured for each seed type was about 600. A total of 6065 images were divided into ten classes, with 50% training, 30% testing, and 20% validation. According to the input requirements of the proposed model, we transformed each image in this dataset to 128 × 128 pixels.

Therefore. This study also aims to produce a new Brassica dataset that didn't already exist and assess how well our CNN architectures perform on it. Digital microscopy was used to capture images of the seeds at 1600 x 1200 pixels and 96 dpi in daylight. The dataset was formed by randomly selecting a large number of images of each class. It was created to test our new CNN architectures and examine the accuracy obtained, which will help us understand how CNN models are used in agricultural data classification tasks. The frequency distribution of the dataset is shown in Table 2.

Data collection and preparation are used in this research's analysis, and it is a task requiring close attention during the analysis process. Therefore, collecting the data is one step in designing and constructing a CNN model. Data preparation for a CNN is the first step of this work. The second step involves data analysis, in which the process includes feature engineering, finding the functional form of the target function, extracting the target function, and finding the functional form of the principal components that can describe all or a large part of the system. In this study, our implementation methodology was based on different procedures. This methodology collects a seed dataset first, followed by model development and revision. Finally, classification is performed, and the performance is evaluated.

Brassica seed datasets are collected from various sources in the first step, and all these data items are then preprocessed to provide practical input to the classifier algorithm. Following data splitting (training, validation, and testing), we used our model for seed class classification, which can be implemented using a learning algorithm. The accuracy of the developed model was then checked, and it was analyzed and evaluated using various performance metrics (see result and discussion section).

**Table 1**
A summary of recent approaches used to classify seeds using deep learning methods.

| Field | Crop | Dataset | Method | Accuaracy | Refference |
|---|---|---|---|---|---|
| Seed classificaiton | grain seeds | grain seeds | LR, CART kNN, SVM, NB, LDA | kNN=87.5% LDA=95.8%, NB=88.05%, LR=91.6% SVM=88.71% CART=88% | Agrawal et al [2] |
| weeds Classification | 8 kinds of weeds | 17508 images | DenseNet | 76% | Foysal et al.[3] |
| Seed classifiaction | 14 kinds of seeds | 2733 images | VGG16 | 99% | Gulzar et al [4] |
| | 5 kinds of seeds | 1250 images | CNN | 87%-89% | Keya et al.[5] |
| | Corn (6 kinds) | 330000 images | MLP, LB, RF BN | MLP: 98.83% LB: 97.78% RF: 97.22% BN: 96.67% | Ali et al [6] |
| | Sugar beet (5 kinds) | 200 images | MSI | 82% | Salimi et al[7] |
| | Brassica rapa | 156 Brassica rapa core collections | New AI models | 87.72% | Minah et al. [8] |
| | Three types of wheat | Wheat dataset | CNN | 84% - 94% | Dubey et al. [9] |

**Table 2**
Description of Brassica seed image dataset.

| ID | Brassica class | Number of Images |
|---|---|---|
| 1 | Brassica Napus Var Annua | 610 |
| 2 | Brassica Napus Var Oleifera | 475 |
| 3 | Brassica Nigra | 653 |
| 4 | Brassica Oleracea Gongylodes | 667 |
| 5 | BrassicaO leraceaLCAV rubra | 650 |
| 6 | Brassica Oleracea Rapa Brassica | 612 |
| 7 | Brassica Oleracea Var Gongylodes | 562 |
| 8 | Brassica Rapa | 562 |
| 9 | Brassica Rapa Oleifera | 494 |
| 10 | Brassica rapa subsp. rapa | 640 |

After implementing the process flow, a more in-depth study of the data is needed to ensure adequate representation of all features for prediction and to improve the decision accuracy of seed class classification. For example, we are dividing the Brassica seed type into ten classes. The implementation of this study is discussed in the following subsection.

### 3.2. Implementation

This subsection focuses on the setup of laboratory experiments for the created Brassica seed classification system using a new model architecture with the TensorFlow framework. Different stages are required to implement the proposed model architectures, beginning with dataset collection and ending with performance evaluation and classification. The classification model is divided into various settings, such as examining the data and building an input workflow to develop a classifier that can predict classes.

We also labeled the data (as shown in Fig. 1) because our new model's learning approach fits into administered learning in deep learning. We have seen that our learning approach can speed up model training and make it possible to improve the quality of the model performance. The primary benefit of the speed and scalability of the model is that it makes it possible to perform faster at a low cost.

In this study, our model has 23 layers including input and output layer for image classification, and each image was used multiple times during the training stage. Through model training, the classification model will experience each training batch exactly once during one epoch and rate its performance on the validation set at the end of each epoch. We tuned hyperparameters like an optimizer, batch size, learning rate, and epoch to implement a model for the dataset. The batch size was adjusted based on the sample size of the dataset, resulting in a batch size of 64. Adam was chosen to be the model architecture's optimizer. The learning rate was adjusted based on learning time, and an appropriate rate of 0.001 was set for the dataset and architectural style. The number of epochs used in this study was 200.

We should observe a decrease in the training and validation loss with each epoch, even if, in practice, the model should be stopped when the loss and accuracy have stabilized. Further, if we can decrease the learning rate by increasing the time step size, training and validation will become less accurate because the additional information gets added to our model. Though a slight change in learning rate might not seem like much, this could have enormous consequences for the success of the model system. Therefore, the weight normalization for the last epoch should be large. It should be adjusted until the loss function becomes stable at the last epoch.

Several extensive experiments were designed to reasonably assess the performance and prove the efficiency of our proposed solution to process seed classification across a wide range of seeds. The experimental results were obtained using a Dell N-series lap-top with a 2.5GHz Intel i6 processor and 8 GB of memory, as well as the new CNN model.

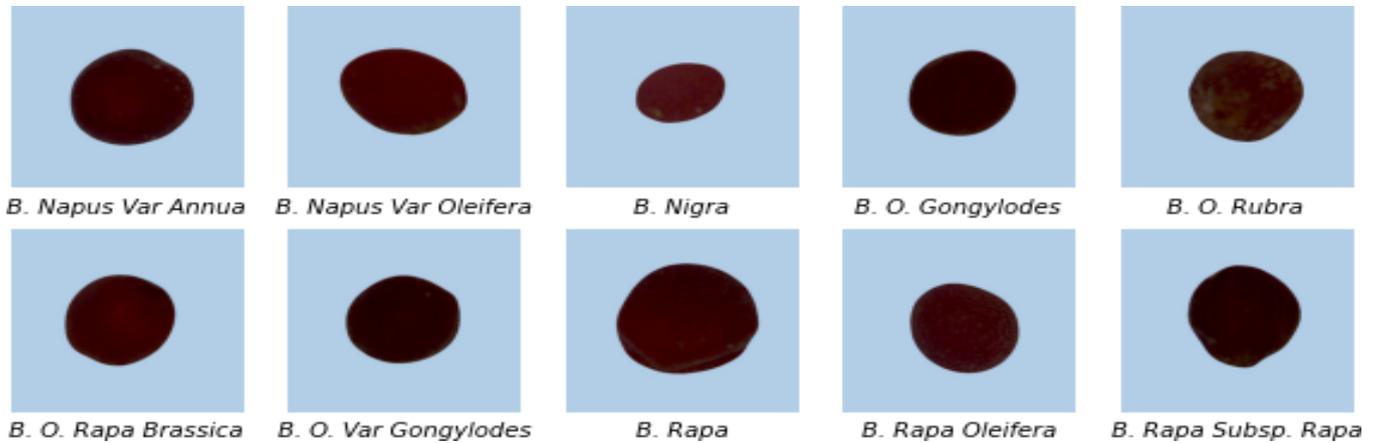

**Fig. 1.** Our Ten-Class Labeled Brassica Dataset

The proposed approach for Brassica type classification is implemented using Python, taking advantage of the powerful and widely adopted open-source TensorFlow library. The entire development and experimentation process was carried out using Google Colab, a cloud-based platform that offers convenient access to computing resources. To expedite the learning process, a personal computer equipped with a GPU was utilized, enabling significant reduction in training time from potentially days to just a few hours. The inclusion of GPU support is crucial in accelerating the processing of multiple examples during each learning iteration, leading to faster convergence and more efficient model training. By harnessing the computational power of GPUs, our approach not only ensures timely experimentation but also facilitates the exploration of larger datasets and more complex architectures, ultimately enhancing the overall performance and reliability of our proposed model for Brassica type classification.

### 3.3. The proposed CNN model architecture

The proposed CNN model for Brassica seed classification consists of 23 layers, including convolutional, pooling, and fully connected layers. The architecture begins with an input layer designed to handle images resized to 128x128 pixels. The initial convolutional layers utilize filters of sizes 5x5 and 3x3 to capture a wide range of features, followed by max pooling layers to reduce the spatial dimensions while retaining essential information.

A unique aspect of our architecture is the use of multiple Conv2D layers with different filter sizes, enhancing the model's ability to capture diverse and complex features. The final layers include two dense layers with 512 neurons each, culminating in a softmax activation function to output the probabilities for the ten seed classes. This structure ensures a balance between depth and computational efficiency, making the model both powerful and scalable. Fig. 2 and Fig. 3 provides a detailed description of the proposed CNN network.

The innovative aspect of this architecture lies in its use of multiple Conv2D layers with varying filter sizes (5x5 and 3x3) and the resultant spatial dimensions of the outputs (42x42, 14x14, 4x4, 2x2). Standard CNN architectures typically employ only one or two Conv2D layers with consistent filter sizes and spatial dimensions. By contrast, our architecture's varied filter sizes and spatial dimensions enable the capture of more diverse and complex features from the input images, thereby enhancing model accuracy. The model culminates with two dense layers of 512 and 10 neurons respectively, and a softmax activation function to output predicted probabilities for each of the ten Brassica types.

Unlike traditional architectures where the output feature maps undergo downsampling through pooling layers, our architecture replaces MaxPooling2D layers with identity mappings. This strategy retains the original spatial dimensions of the feature maps, allowing the network to capture finer spatial information at each stage. Preserving spatial resolution throughout the network offers a novel perspective and potentially enables the model to capture intricate patterns specific to Brassica seed classification tasks.

Our network's convolutional layer, with a receptive field sufficiently large for image feature extraction, facilitates the identification of subtle features necessary for accurate Brassica seed classification. The relevant network depth was also optimized to reduce the model size while maintaining high accuracy. These improvements in convolution layers are crucial for enhancing the model's classification performance. Additionally, we compared our model's performance against three well-known state-of-the-art CNN models: InceptionV3, DenseNet121, and ResNet152. Each model was evaluated using a global average pooling layer followed by a flatten layer, a fully connected layer, and a softmax layer with ten outputs per epoch. We also assessed susceptibility to overfitting by tuning the number of hidden units in each network.

By comparing the performance of different CNN models, we gained insights into the impact of architectural variations on their effectiveness. Traditional approaches, such as feature engineering and dimensionality reduction, have been extensively utilized. However, as deep learning methods like CNNs advance, they offer more efficient alternatives.

As depicted in Fig. 4, we present the network diagrams of the three CNN models under investigation. These diagrams offer a visual representation of the architectural structure, illustrating the flow of information through the layers and highlighting the distinctive characteristics of each model. By studying these network diagrams, we can better understand the design choices made in each model and their potential impact on performance. These models were carefully selected to represent different architectural designs, enabling us to assess their respective strengths and weaknesses. By thoroughly examining their performance metrics, including accuracy, loss, and computational efficiency, we gained valuable insights into the effectiveness of specific architectural choices. The outcomes of our analysis not only contribute to the current body of research on CNN models but also shed light on promising avenues for future investigations. The continuous de-velopment of CNN architectures and techniques holds great potential for advancing various computer vision tasks and applications. Furthermore, the comparison of these models serves as a foundation for our research, allowing us to identify the most effective model for our specific task of Brassica type classification. Additionally, it provides insights into the broader field of CNN model design, offering valuable knowledge that can guide future research and advancements in the field.

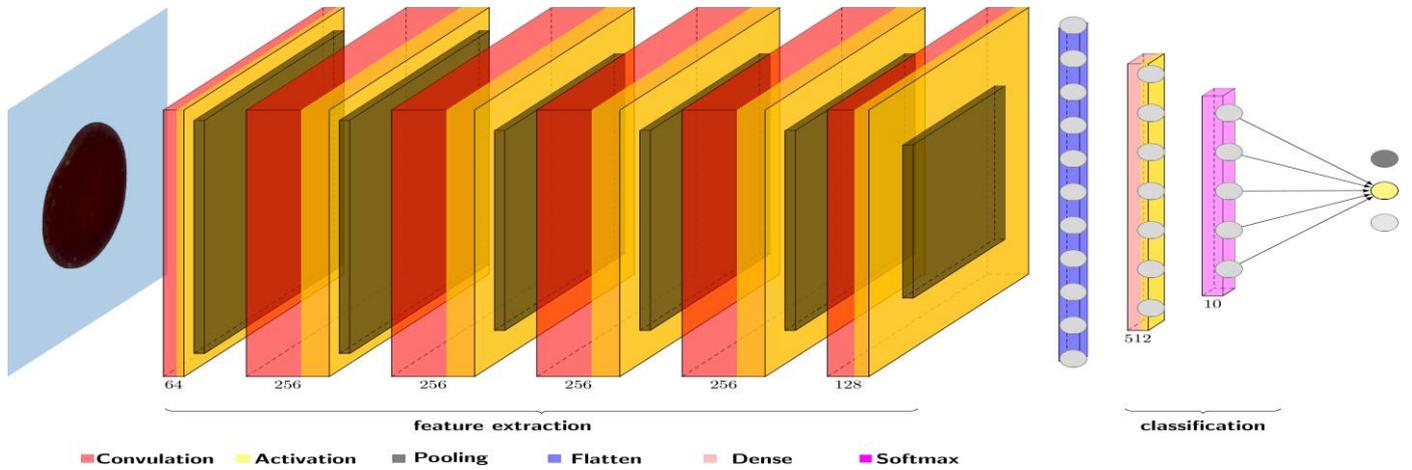

**Fig. 2.** Proposed CNN model for brassica seeds classification.

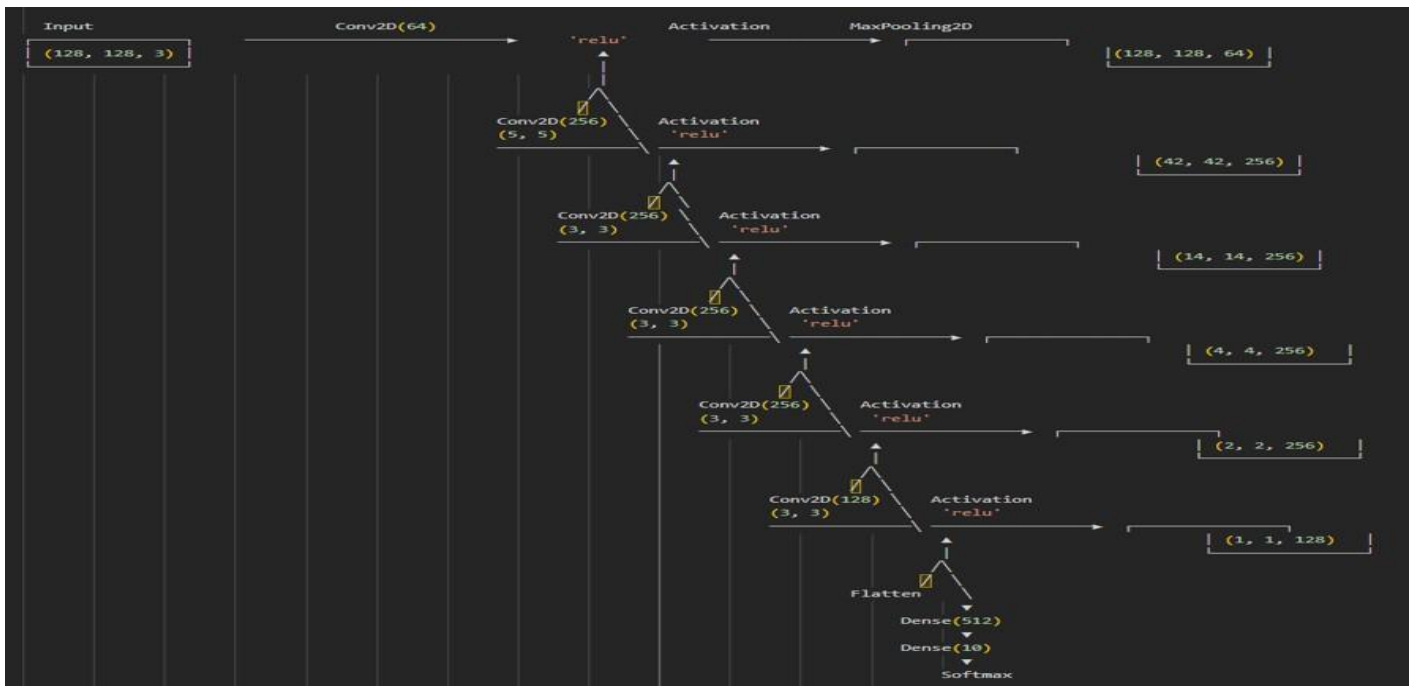

**Fig. 3.** A detailed visual depiction of the proposed architecture

We conducted a comprehensive analysis and comparison of the performance of different models using a carefully collected Brassica seed image dataset. To ensure fair evaluation, we employed the baseline training approach, where all layers of the models were trained using our Brassica seed dataset. Additionally, we utilized pre-trained weights that were randomly initialized during the training process for each of the three models under investigation.

The training and validation datasets were subjected to rigorous experimentation for a total of 200 epochs, employing a batch size of 64. This extensive training duration allowed us to capture the nuances and intricacies of the dataset and evaluate the models' learning capabilities over an extended period. In the forthcoming sections, we present a detailed analysis of the results obtained during both the training and validation phases, shedding light on the models' performance and their ability to generalize to unseen data.

To facilitate the training process, we employed the Adam optimizer along with carefully selected hyperparameters, including a learning rate of 0.001. The choice of hyperparameters greatly influences the training dynamics and convergence of the models, and our selection aimed to strike a balance between stability and rapid learning. During the training process, the categorical cross entropy function served as the loss function, enabling us to effectively measure the dissimilarity between predicted and actual class labels. Moreover, the SoftMax activation function was employed, enabling the models to generate probability distributions over the target classes.

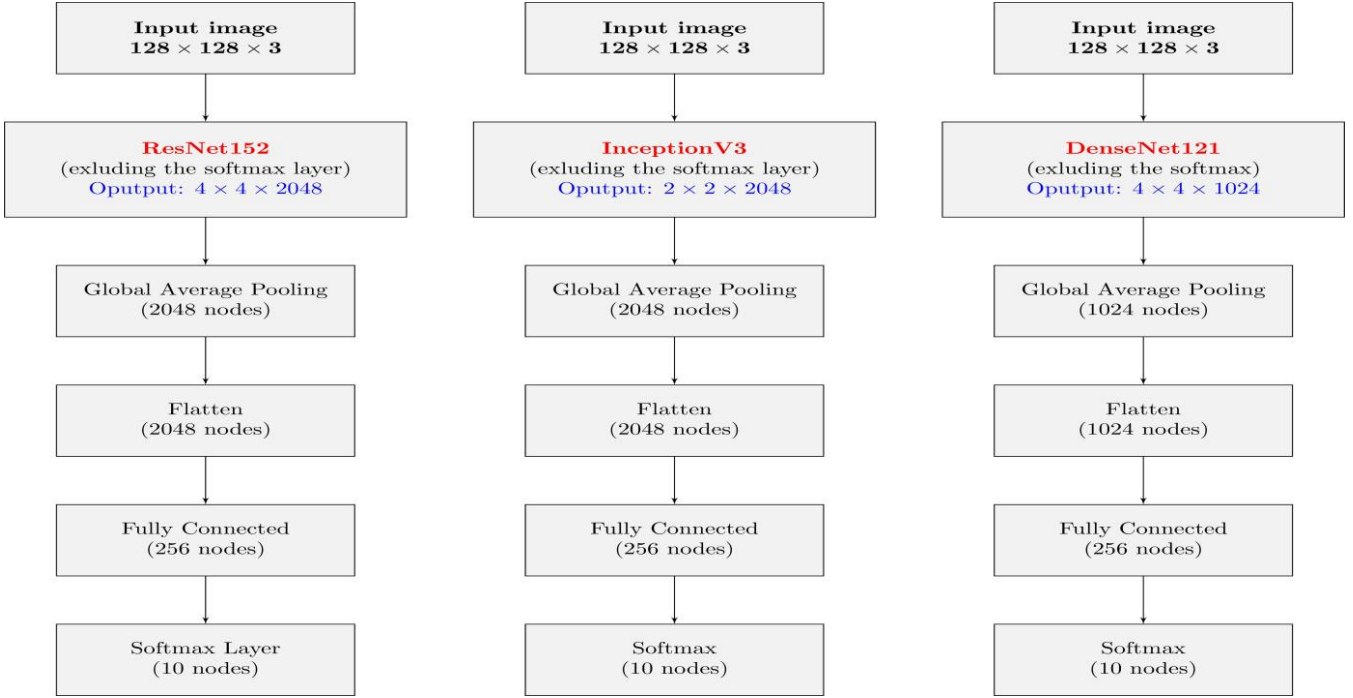

**Fig. 4.** Architecture diagrams of the models used.

In the subsequent sections, we delve deeper into the details of the three models under examination and provide a comprehensive overview of our proposed model. We discuss their architectural nuances, highlight their strengths and weaknesses, and present empirical evidence supporting the efficacy of our novel approach.

## 4. Result and Discussion

In this section, we conducted several experiments to evaluate and demonstrate the performance of the proposed approach on the Brassica seed dataset. Section 4.1, Section 4.2, and Section 4.3 provide a detailed explanation of the results obtained.

### 4.1. Effects of Batch size and Epoch selection on model training: Experimental Findings

In the pursuit of optimal model training, the selection of batch size and epoch values plays a crucial role. In this section, we introduce the results of our comprehensive experiments conducted to determine the impact of different batch sizes and epochs on model performance. The findings shed light on the optimal parameters for training neural networks, providing valuable insights for researchers and practitioners alike.

Fig. 5 showcases the outcomes of our proposed model trained using batch sizes of 8, 16, 32, and 64. The training time per epoch and testing accuracy were examined as batch sizes increased. Remarkably, Fig. 5'a and Fig. 5'b exhibit a compelling trend: as batch sizes increased, the training time per epoch decreased while the testing accuracy soared. This observation underscores the significance of selecting an appropriate batch size to achieve optimal results.

Further analysis revealed that a batch size of 64 yielded the most effective outcomes throughout the model training process. The testing accuracies at various model training epochs were evaluated and illustrated in Fig. 6. Notably, the testing accuracy displayed a gradual improvement with the progression of epochs, ultimately reaching a peak at 200 iterations. Therefore, 200 epochs were chosen as a balance between training time and accuracy, indicating the potential of this configuration for achieving commendable performance.

While the results obtained from our study provide valuable insights into selecting optimal parameters for training neural networks, it is crucial to acknowledge that these findings are specific to the dataset and model architecture utilized. Different datasets and model structures may exhibit variations in optimal parameter settings. Thus, caution must be exercised when generalizing these findings to other models and datasets.

To comprehensively explore the landscape of optimal parameter selection, it is highly recommended to replicate similar experiments using different architectures and diverse datasets. Such endeavors will allow researchers to identify the best-suited parameters for each specific case, contributing to the continuous advancement of model training practices.

The effects of batch size and epoch selection on model training have been thoroughly investigated in this study. The experimental findings demonstrate the intricate relationship between these parameters and model performance. By highlighting the advantages of utilizing a batch size of 64 and training for 200 epochs, this research provides practical insights for achieving superior results in neural network training. However, it is essential to exercise caution in generalizing these findings and adapt the parameters to the specific requirements of each dataset and model architecture. Future research endeavors should expand upon these findings and explore parameter selection in diverse contexts, fostering advancements in model training techniques.

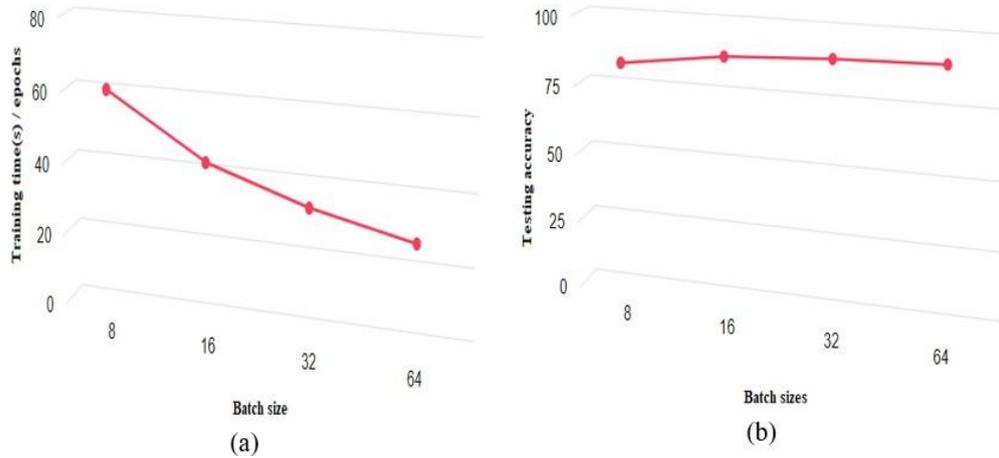

**Fig. 5.** The effect of batch sizes on the model's performance is shown in (a) batch size vs. training time per epoch and (b) batch size vs. the model's testing accuracy.

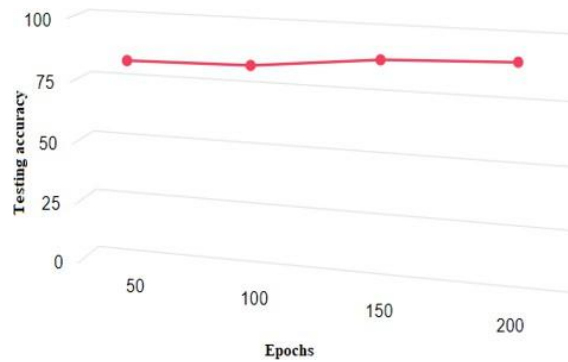

**Fig. 6.** Effect of epochs on testing accuracy.

*4.2. Application of the proposed model architecture*

Brassica seed classification holds paramount importance in the field of cultivation. With the aim of developing an automated seed classification model, conducting various experiments to test novel deep learning models becomes an indispensable task, offering numerous benefits. In light of this, we present the application of our proposed model architecture in a Brassica seed classification test experiment using the collected dataset.

To adapt the model to the Brassica seed dataset, several adjustments were made. Softmax activation was employed in the model's output layer to enable it to output probabilities for ten distinct seed classes. Additionally, hyperparameter settings were fine-tuned before commencing the training process. The model was programmed to run for 200 epochs, allowing it to learn and refine its representations over an extended period. During training, a learning rate of 0.001 was utilized, considering the crucial role that learning rate plays in influencing model performance and preventing overfitting.

In addition to learning rate optimization, other crucial hyperparameters, such as the optimizer and batch size, were carefully tuned to maximize model effectiveness. Considering the sample size of the dataset, the batch size was set to 64, striking a balance between computational efficiency and model performance. The Adam optimizer, known for its efficacy with large-scale datasets, was chosen as the optimization algorithm for our model architecture. The meticulous adjustment of hyperparameters is pivotal to ensure the model's ability to generalize well across diverse Brassica seed samples.

The performance of our model was rigorously evaluated using various metrics. The accuracy curve was employed to assess the model's predictive capability and its capacity to accurately classify Brassica seed types within 200 epochs (see Fig. 7). The results unequivocally demonstrated the accuracy and reliability of our proposed models.

The evaluation metrics used, including average training accuracy of 96.10%, average validation accuracy of 95.47%, training loss of 0.3478, and validation loss of 0.4390, highlight the exceptional performance of the proposed architecture. Furthermore, the accuracy and loss graphics, depicted in Fig. 7, visually reinforce the model's prowess in accurately classifying Brassica seed types.

A crucial factor contributing to the model's robust performance lies in the pre-processing techniques employed. The stability observed during the training and validation processes can be attributed to the proposed architecture, meticulous data collection, and the strategic distribution of data across all classes. Moreover, the incorporation of the dropout technique played a pivotal role in enhancing the model's validation performance, ensuring it did not deviate significantly from its training performance.

Overall, the application of our proposed model architecture in Brassica seed classification exemplifies its effectiveness and accuracy. The comprehensive adjustments made to hyperparameters, coupled with the utilization of pre-processing techniques, yielded a stable and high-performing model. The results obtained serve as a testament to the model's capability to accurately classify Bras-sica seed types and underscore its potential for automating seed classification tasks.

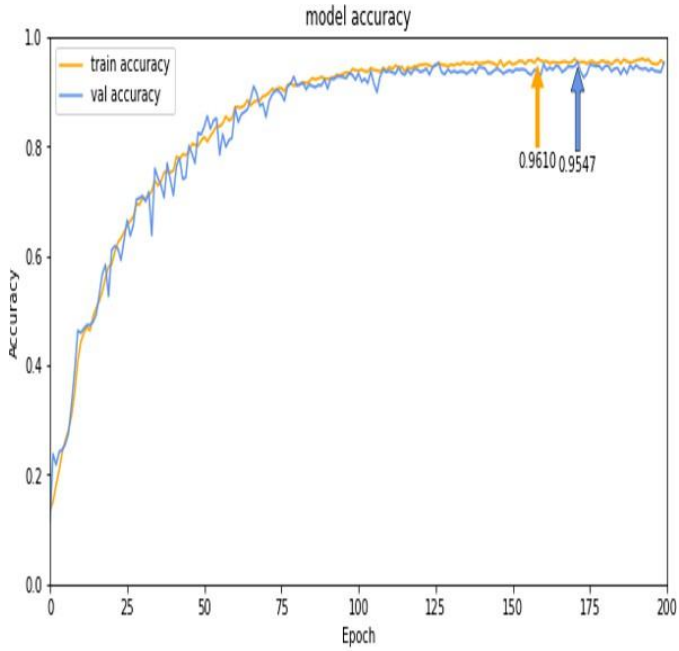 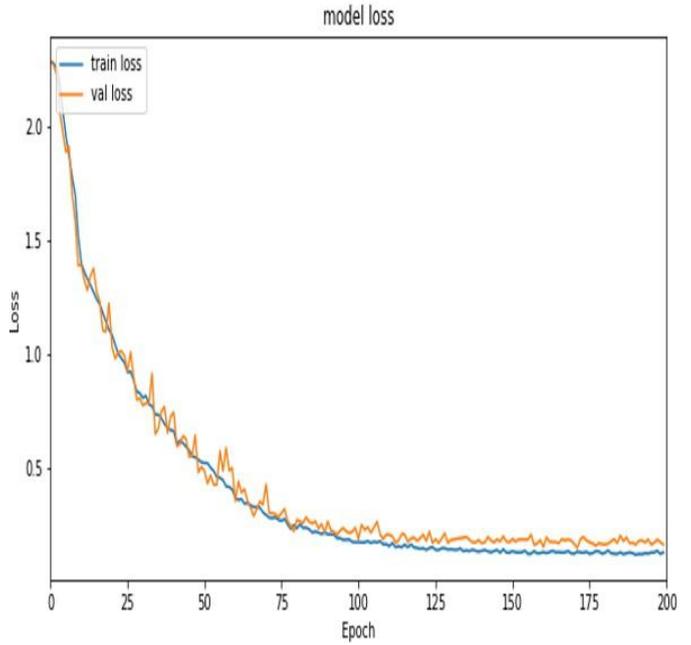

**Fig. 7.** Training and validation accuracy and loss of the proposed model.

### 4.2.1. Performance evaluation of the proposed model for image classification of Brassica seeds classes

In this paper, we used the confusion matrix to provide a clear insight into the accuracy and ways in which our classification model is confused when making predictions. In this work, the confusion matrix has four metrics, each of which measures the accuracy of classifications and attempts to gauge how each pair of predictor and target attributes will behave for one given class value. Therefore, the confusion matrix was employed to visualize the effectiveness of the CNN model. This data on the confusion matrix represents the true class in the samples as well as the class predicted by the CNN classifier.

Thus, using these two labeled sets, we summarized the results of testing the classifier that distinguishes between ten classes of Brassica seed. The four metrics were commonly true positives (TP), true negatives (TN), false positives (FP), and false negatives (FN). In this study, TP and TN represent correct identification of Brassica seeds, while FP and FN represent incorrect identification. The confusion matrices for the models have been depicted in Fig. 8.

The proposed method and architecture trained with the Brassica dataset image indicated that the proposed CNN model was good at predicting the image of ten classes. Therefore, our technique's analysis, evaluation, and validation tasks were carried out. The results were satisfactory for the proposed CNN model used to classify the image of ten classes, as shown in Fig. 8. The method resulted in an accuracy of 93% on the test dataset, which included 1214 images merged from 10 different classes. Furthermore, the model has achieved 95.56% training accuracy and 94.21% validation accuracy of Brassica seed classifications after 200 epochs, as shown in Fig. 9.

The proposed model's performance was evaluated using some statistical parameters of the confusion matrix, such as accuracy, precision, recall, and the F1-score. These performance measures were selected because they were the most commonly used metrics in previous studies to evaluate the performance of most methods [18]. Therefore, the performance evaluation equations in Eqs:

(1), (2), (3), and (4) are used to calculate performance measures and evaluate results.

$$Accuracy = \frac{TP + TN}{TP + TN + FP + FN} \quad (1)$$

**Fig. 8.** Confusion matrices of the proposed models on Brassica seeds dataset.

```
              precision    recall  f1-score   support

       Annua     0.9746    0.9829    0.9787       117
 Var Oleifera    0.9211    0.9633    0.9417       109
       Nigra     0.9847    0.9699    0.9773       133
   Gongylodes    0.9124    0.9843    0.9470       127
        Rubra    0.9375    0.8974    0.9170       117
 Rapa Brassica   0.9314    0.7540    0.8333       126
 VarGongylodes   0.9400    0.9156    0.9276       154
         Rapa    0.7769    0.9352    0.8487       108
  Rapa Oleifera  0.9667    0.9255    0.9457        94
   Subsp Rapa   0.9692    0.9767    0.9730       129

     accuracy                        0.9300      1214
    macro avg    0.9314    0.9305    0.9290      1214
 weighted avg    0.9332    0.9300    0.9297      1214

val_acc     0.9421    std 0.0052
train_acc:  0.9556    std 0.0039
```

**Fig. 9.** The overall performance of the proposed model.

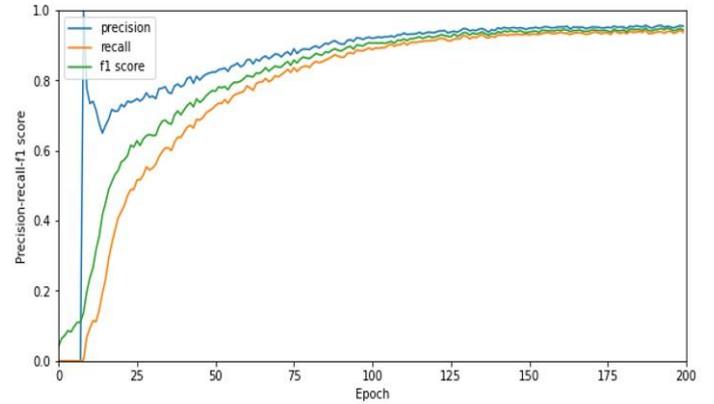

$$Sensitivity(Recall) = \frac{TP}{TP + FN} \quad (2)$$

$$F1 - Score = \frac{Precision \times Recall}{Precision + Recall} \quad (3)$$

$$Precision = \frac{TP}{TP + FP} \quad (4)$$

where

- $TP$ = True Positive
- $TN$ = True Negative
- $FP$ = False Positive
- $FN$ = False Negative

Since we have all the necessary metrics for all the classes from the confusion matrix, we calculated the performance measures for these classes, as shown in Fig. 9 and Fig. 10. As can be seen, the proposed method yielded an accuracy of 93%, 95.56% training accuracy, and 94.21% validation accuracy; Moreover, looking at the accuracy obtained by the model on this collected dataset, it is safe to say that the model is doing well regardless of the dataset it is evaluated on (i.e., collected dataset) as the achieved accuracy is 93% on this dataset.

Fig. 9 illustrates the performance of each seed class mentioned in the proposed model in terms of recall, precision, F1 score, and support [19, 20]. The number of instances of each class that were performed during model training is represented by support, and the ratio of correctly estimated samples in the model to the total dataset is represented by accuracy. It is important to note that, as shown in Fig. 9, the model achieved the highest possible values for each seed class and in all metrics (precision, recall, F1-score) in both the training and validation sets except the recall of class 5 and the precision of class 7, this is because the model mixed up these two classes due to their texture similarities. This could have been caused by the camera's light settings. Fig. 9 depicts the overall performance measure of the proposed model, and Fig. 10 shows a curve illustrating how well the performance metric performs in both training and validation.

### 4.3. Comparaison of the proposed CNN model to some pre-trained state-of-the-art deep learning methods.

In this study, we analyzed and evaluated the performance of our CNN model by providing a comparative analysis of the classification performance of the proposed model with that of pre-trained models. Therefore, we presented close results between the pre-trained models and our proposed model, as shown in Table 3. The parameters and the architectures are given in Fig. 4 were selected for transfer learning. The prepared Brassica dataset was trained using Inception-v3, Densnet121, and Resnet152. In addition, the optimal parameters in Fig. 4 were used to prevent over-fitting during training and avoid spending more time. All networks have been trained for 200 epochs. The classification results for all varieties of Brassica seeds in the different models are shown in Table 3.

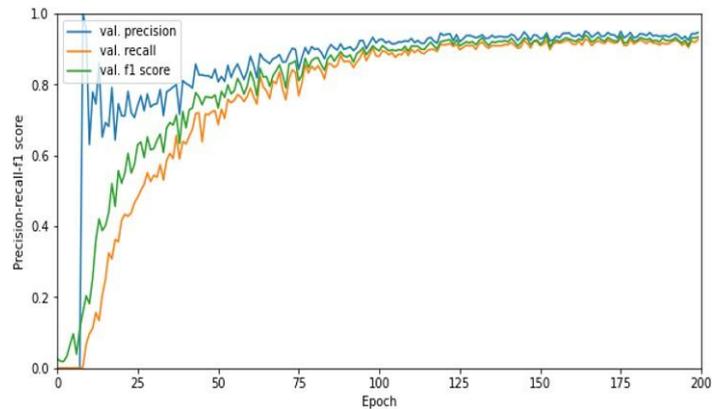

**Fig. 10.** The training and validation results of the proposed model.

It is apparent from Table 3 that among pre-trained models, Densent121 achieved the highest accuracy of 90.03%, Inceptionv3 achieved 84.71%, and Resnet152 achieved the least (73.34%). The results show that the highest-performing Densent121 model is behind the proposed model in classification accuracy, with an accuracy of 93%. Therefore, it is evident that our model's performance in terms of accuracy, average precision, recall, and f1-score was significantly better than that of pre-trained models. For example, Densent121 reported the average precision, recall, and f1-score of 92.45%, 90.03, and 90.11%, respectively, while our model reported 90.78%, 93.30%, and 90.26% for these metrics.

These results reported that it is possible to achieve satisfactory and better classification performances in proposed model training than in pre-trained ones and improve the performance to achieve better accuracy in image classification. Furthermore, it suggests that low-level and high-level features can be successfully extracted from the image dataset under study using the suggested learning approaches. One of its advantages is the proposed model's ability to process large amounts of data more easily than other deep learning methods. Additionally, the suggested methods can use a combination of transfer learning and feature selection to improve

**Table 3**
Overall performance of CNN architectures.

| Method | accuracy | precision | recall | F1 score |
| --- | --- | --- | --- | --- |
| Our model | 0.930 | 0.9078 | 0.930 | 0.9026 |
| Resnet152 | 0.7334 | 0.8613 | 0.7334 | 0.7279 |
| Inceptionv3 | 0.8471 | 0.8745 | 0.8471 | 0.8212 |
| DenseNet121 | 0.9003 | 0.9245 | 0.9003 | 0.9011 |

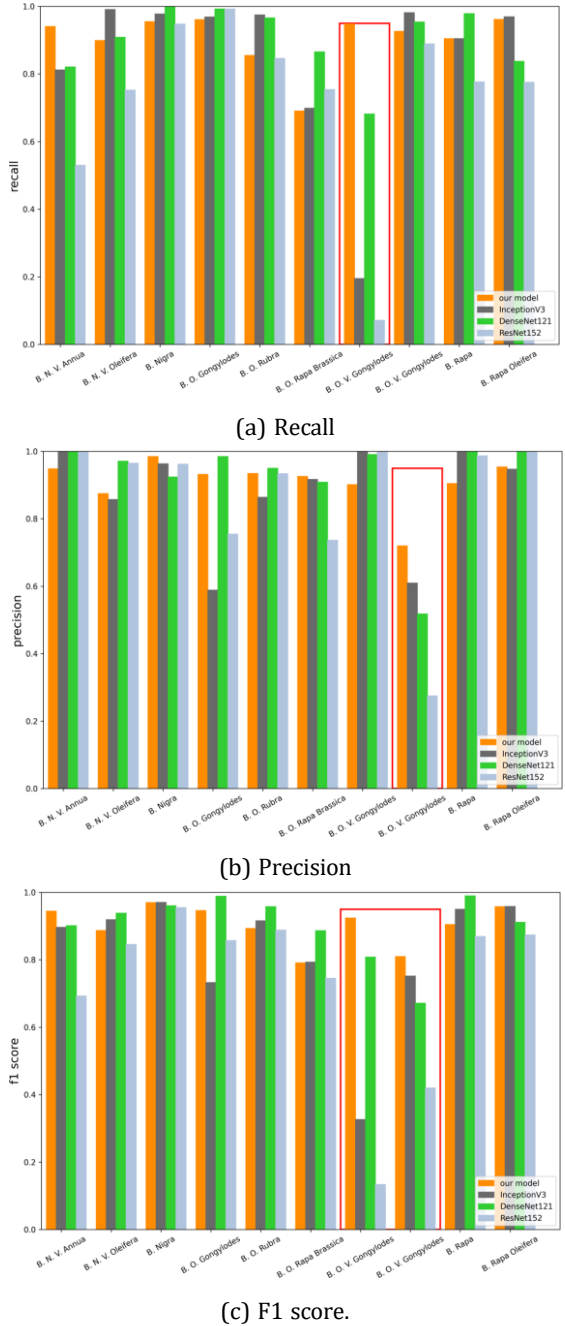

(a) Recall

(b) Precision

(c) F1 score.

**Fig. 11.** Comparative analysis of classification performance based on performance metrices: a) recall, b) precision, c) f1 score.

performance.

Next, in addition to the model accuracy, we used class accuracy, which seems more descriptive. We analyzed the performance of each seed class mentioned in the proposed model in terms of precision, recall, and F1-score, as shown in Fig. 11, which provide a comparative analysis of the classification performance of the proposed model with pre-trained models. Fig. 11 shows that the models' values are close in for all metrics (precision, recall, and F1-score), and all the models showed the maximum possible values for each seed class except the seed classes motioned with the red line. It is so clear that there is a significant difference in the model's performance when it comes to the red line; classes with a red line show that all models' performances become confused and unstable, except for our model, which maintained good performance across all classes. Again, this proves how stable and accurate our model is compared to others. Fig. 11 displays the classification results for all varieties of Brassica seeds using various models.

This statistic demonstrates how our model's overall performance remains stable. As shown in Fig. 11, the classification accuracy of four different models for ten Brassica seeds was controllable and convergent in all classes until the classes were indicated with a red line, resulting in poor distinction. Our model's accuracy was over 90%, and Densnet121 reported the second-best performance. These results indicated that the Inception-v3, Densnet121, and Resnet152 models needed to be better adapted to these varieties, and this might be due to texture similarities. On the other hand, our model classification results were still very encouraging and showed the maximum possible values for each seed class. Therefore, the experimental analysis revealed that our model, compared with pretrained models' architectures with updated weights and fine-tuning, had good generalization capability in the Brassica seed dataset. The Brassica dataset was widely used as an example of a high-level feature dataset for these models' training and was used to assess the robustness of the proposed method. The results indicate that our model performed exceptionally well in terms of generalization on this dataset.

The proposed model exhibits superior performance and accuracy compared to other networks mentioned in the literature. The observed difference in classification accuracy surpasses 2%, further reinforcing the model's superiority. These findings validate the practicality and feasibility of developing a new model and designing a network that can effectively address challenging classification tasks. Moreover, the unique structure of the proposed network successfully combines depth and width, resulting in an optimal model for image classification. This combination of depth and width allows for the creation of a network architecture that is not only powerful but also natural in its design. By achieving such optimal performance, the proposed model offers practical solutions for complex classification tasks and holds great potential for various applications in different fields.

In summary, the proposed model surpasses existing networks in terms of performance and accuracy, making it a reliable and practical choice. Its ability to effectively merge depth and width provides a valuable solution for challenging classification tasks and sets the stage for future applications in diverse fields.

## 5. Conclusion

Deep learning in agriculture continues to make significant contributions in many areas. As a result, using deep learning tech-niques, the field of agriculture is rapidly advancing in a way that will definitely change it for the better. Therefore, studies on the deployment of Deep learning techniques for seed classification are yielding promising results, with the potential to alleviate concerns about food security by providing a cost-effective alternative.

This work focused on developing and proposing a new CNN model for multiclass Brassica seed image classification tasks. Furthermore, we evaluated the performance of various CNNs and compared them to our CNN model for this task. The goal was to evaluate the feasibility of determining the best-performing model's architecture and the best training options for this problem. The goal of this study is also to create a new Brassica dataset that did not previously exist and evaluate the performance of our CNN architectures on this dataset. We also compared the proposed approach to a series of pre-trained models, including Densent121, Inceptionv3, and Resnet152, demonstrating that our proposed model could significantly improve the accuracy of CNNs in predicting expression values. Finally, using various measurements, we evaluated the effects of the proposed architectures and training settings on performance improvement. The proposed approach's results showed our model accuracy of up to 93%. On the other hand, Densnet121 reported 90.03%, Inceptionv3 achieved 84.71%, and Resnet152 achieved the least (73.34%).

The results obtained during this study can serve as inspiration for other similar visual object recognition, so the practical study of this work will easily extend to the classification problems of other seeds images. Our proposed model for Brassica seeds classification was successfully implemented, discussed, and a satisfactory classification result was obtained. The model has been fast and accurate, but it has only been tested for Brassica classification rather than for other scenarios. As a result, this work will be expanded to work on different datasets with dissimilar seeds in the future.

**Declarations**

- Funding: This research received no external funding.
- Conflict of interest: The authors declare no conflict of interest.
- Informed Consent Statement: Not applicable.

**References**

[1] A.S. Universit., seed collection and plant genetic diversity, https://embryo.asu.edu/pages/seed-collection-and-plant-genetic-diversity-1900-1979.

[2] D. Agrawal, P. Dahiya, Comparisons of classification algorithms on seeds dataset using machine learning algorithm, Compusoft 7 (5) (2018) 2760–2765.

[3] F.A. Foysal, M. Shakirul Islam, S. Abujar, S. Akhter Hossain, et al., A novel approach for tomato diseases classification based on deep convolutional neural networks, in: Proceedings of International Joint Conference on Computational Intelligence, Springer, 2020, pp. 583–591. URL: https://doi.org/10.1007/978-981-13-7564-4_49.

[4] Y. Gulzar, Y. Hamid, A.B. Soomro, A.A. Alwan, L. Journaux, A convolution neural network-based seed classification system, Symmetry 12 (12) (2020) 2018. URL: https://www.mdpi.com/2073-8994/12/12/2018.

[5] M. Keya, B. Majumdar, M.S. Islam, A robust deep learning segmentation and identification approach of different bangladeshi plant seeds using cnn, in: 2020 11th International Conference on Computing, Communication and Networking Technologies (ICCCNT), IEEE, 2020, pp. 1–6. URL: 10.1109/ICCCNT49239.2020.9225677.

[6] A. Ali, S. Qadri, W.K. Mashwani, S. Brahim Belhaouari, S. Naeem, S. Rafique, F. Jamal, C. Chesneau, S. Anam, Machine learning approach for the classification of corn seed using hybrid features, International Journal of Food Properties 23 (1) (2020) 1110–1124. URL: 10.1080/10942912.2020.1778724.

[7] Z. Salimi, B. Boelt, Classification of processing damage in sugar beet (beta vulgaris) seeds by multispectral image analysis, Sensors 19 (10) (2019) 2360. URL: https://doi.org/10.3390/s19102360.

[8] M. Jung, J.S. Song, S. Hong, S. Kim, S. Go, Y.P. Lim, J. Park, S.G. Park, Y.M. Kim, Deep learning algorithms correctly classify brassica rapa varieties using digital images, Frontiers in Plant Science 12. URL: DOI:10.3389/fpls.2021.738685.

[9] B. Dubey, S. Bhagwat, S. Shouche, J. Sainis, Potential of artificial neural networks in varietal identification using morphometry of wheat grains, Biosystems engineering 95 (1) (2006) 61–67. URL: DOI:10.1016/j.biosystemseng.2006.06.001.

[10] F. Guevara-Hernandez, J.G. Gil, A machine vision system for classification of wheat and barley grain kernels, Spanish Journal of Agricultural Research (3) (2011) 672–680. URL: DOI:10.5424/sjar/20110903-140-10.

[11] M. Shahid, M.S. Naweed, S. Qadri, Mutiullah, E.A. Rehmani, Varietal discrimination of wheat seeds by machine vision approach, Life Sci 11 (6) (2014) 245–252.

[12] A. Loddo, M. Loddo, C. Di Ruberto, A novel deep learning based approach for seed image classification and retrieval, Computers and Electronics in Agriculture 187 (2021) 106269. URL: https://doi.org/10.1016/j.compag.2021.106269.

[13] V. Maeda-Gutiérrez, C.E. Galvan-Tejada, L.A. Zanella-Calzada, J.M. Celaya-Padilla, J.I. Galván-Tejada, H. Gamboa-Rosales, H. Luna-Garcia, R. Magallanes-Quintanar, C.A. Guerrero Mendez, C.A. Olvera-Olvera, Comparison of convolutional neural network architectures for classification of tomato plant diseases, Applied Sciences 10 (4) (2020) 1245. URL: DOI:10.3390/app10041245.

[14] C. Szegedy, W. Liu, Y. Jia, P. Sermanet, S.E. Reed, D. Anguelov, D. Erhan, V. Vanhoucke, A. Rabinovich, Going deeper with convolutions, 2015 IEEE Conference on Computer Vision and Pattern Recognition (CVPR) (2014) 1–9. URL: https://doi.org/10.48550/arXiv.1409.4842.

[15] A. Krizhevsky, I. Sutskever, G.E. Hinton, Imagenet classification with deep convolutional neural networks, Communications of the ACM 60 (6) (2017) 84–90. URL: DOI:10.3390/app10041245.

[16] C. Szegedy, V. Vanhoucke, S. Ioffe, J. Shlens, Z. Wojna, Rethinking the inception architecture for computer vision, 2016 IEEE Conference on Computer Vision and Pattern Recognition (CVPR) (2015) 2818–2826. URL: https://doi.org/10.48550/arXiv.1512.00567.

[17] K. He, X. Zhang, S. Ren, J. Sun, Deep residual learning for image recognition, 2016 IEEE Conference on Computer Vision and Pattern Recognition (CVPR) (2015) 770–778. URL: DOI:10.1109/CVPR.2016.907.

[18] S. Abed, A.A. Esmaeel, A novel approach to classify and detect bean diseases based on image processing, 2018 IEEE Symposium on Computer Applications & Industrial Electronics (ISCAIE) (2018) 297–302. URL: DOI:10.1109/ISCAIE.2018.8405488.

[19] D.M.W. Powers, Evaluation: from precision, recall and f-measure to roc, informedness, markedness and correlation, ArXiv abs/2010.16061. URL: https://doi.org/10.48550/arXiv.2010.16061.

[20] H. Alatrista-Salas, J. Morzán-Samamé, M.N. del Prado, Crime alert! crime typification in news based on text mining, Springer Verlag, Alemania, 2020, pp. 725–741. URL: doi="10.1007/978-3-030-12388-8_50".